\pdfoutput=1

\documentclass[11pt]{article}

\usepackage[final]{acl}

\usepackage{times}
\usepackage{latexsym}
\usepackage{booktabs}
\usepackage{multirow}
\usepackage{amsmath}
\usepackage{amssymb}
\usepackage{adjustbox}
\usepackage{enumitem}
\usepackage{placeins} 
\usepackage{amsmath}
\usepackage{makecell}
\usepackage{xcolor}
\usepackage{booktabs}
\usepackage[mathscr]{eucal}
\usepackage{xcolor} 
\usepackage{tabularx}
\usepackage{hyperref}
\usepackage{graphicx}
\usepackage{ragged2e} 

\usepackage[T1]{fontenc}

\usepackage[utf8]{inputenc}

\usepackage{microtype}

\usepackage{inconsolata}

\usepackage{graphicx}
\usepackage{xspace}
\renewcommand{\vec}[1]{\mathbf{#1}}
\newcommand{\ours}{CIEA\xspace}

\title{Enhancing Multimodal Retrieval via Complementary Information Extraction and Alignment}

\author{
    Delong Zeng\textsuperscript{\rm 1,}\thanks{Work done as an intern at Alibaba Group.}\;
    Yuexiang Xie\textsuperscript{\rm 2}\;
    Yaliang Li\textsuperscript{\rm 2}\;
    Ying Shen\textsuperscript{\rm 1,3,}\thanks{Corresponding author.} \\
    \textsuperscript{\rm 1}School of Intelligent Systems Engineering, Sun Yat-sen University \textsuperscript{\rm 2}Alibaba Group\\
    \textsuperscript{\rm 3}Guangdong Provincial Key Laboratory of Fire Science and Intelligent Emergency Technology\\
    \texttt{zengdlong@mail2.sysu.edu.cn, \{yuexiang.xyx, yaliang.li\}@alibaba-inc.com}\\\texttt{sheny76@mail.sysu.edu.cn}
}

\begin{document}
\maketitle
\begin{abstract}
Multimodal retrieval has emerged as a promising yet challenging research direction in recent years. Most existing studies in multimodal retrieval focus on capturing information in multimodal data that is similar to their paired texts, but often ignore the complementary information contained in multimodal data. In this study, we propose \ours, a novel multimodal retrieval approach that employs {\bf C}omplementary {\bf I}nformation {\bf E}xtraction and {\bf A}lignment, which transforms both text and images in documents into a unified latent space and features a complementary information extractor designed to identify and preserve differences in the image representations. We optimize \ours using two complementary contrastive losses to ensure semantic integrity and effectively capture the complementary information contained in images.
Extensive experiments demonstrate the effectiveness of \ours, which achieves significant improvements over both divide-and-conquer models and universal dense retrieval models. We provide an ablation study, further discussions, and case studies to highlight the advancements achieved by \ours.  
To promote further research in the community, we have released the source code at \href{https://github.com/zengdlong/CIEA}{https://github.com/zengdlong/CIEA}.
\end{abstract}

\section{Introduction}

Retrieval-Augmented Generation (RAG)~\cite{lewis2020retrieval, gao2023retrieval, yi2024survey} has recently attracted widespread attention for its role in enhancing large language models (LLMs)~\cite{touvron2023llama,bai2023qwen,chung2024scaling,brown2020language} by providing up-to-date information and alleviating hallucination issues. 
Most existing studies focus on retrieving textual information~\cite{cheng2024lift,shi2024replug,wang2024searching,zhu2025knowledge}, leveraging text corpora to provide factual support for model responses. As the scale of multimodal data continues to grow, effectively supporting multimodal retrieval to provide knowledge beyond text has emerged as a promising yet challenging research area~\cite{zhao2023retrieving,kuang2024natural}.

\begin{figure}[!t]
    \centering
    \includegraphics[width=\linewidth]{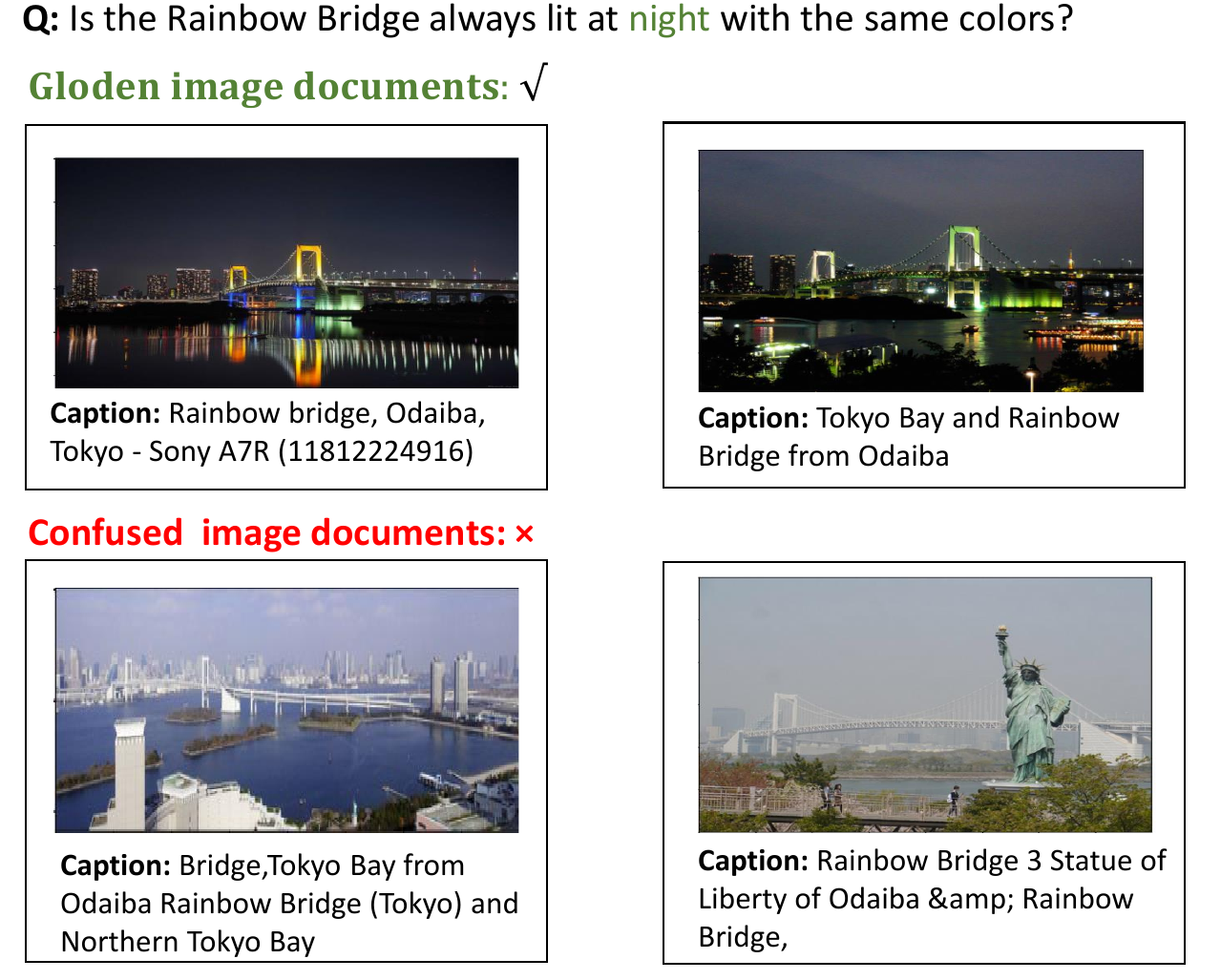}
    \caption{An example of multimodal retrieval. The images contain critical information that can assist in query responses, which might not be present in the text.}
    \label{fig:rag-example}
\end{figure}

To promote the progress of multimodal retrieval, researchers have employed captioning models to convert multimodal data into text~\cite{wu2023cap4video,baldrati2023zero,mahmoud2024sieve}, enabling the application of existing retrieval techniques developed for text corpora. However, these approaches can be heavily influenced by the effectiveness of the captioning models, which might result in the loss of crucial information during the conversion process~\cite{che2023enhancing, li2024monkey}. Recent studies propose representation-based approaches, wherein textual and multimodal information are mapped into a unified embedding space for knowledge retrieval, after a separate encoding process~\cite{clip,zhang2021vinvl} or a project-based joint encoding process~\cite{zhou2024marvel, li2024improving}.

In this study, we focus on a novel representation-based approach that aims to transfer multimodal data into the latent vector space of language models. We identify a critical limitation of existing representation-based methods~\cite{wang2024unified,zhou2024marvel}: they tend to focus on capturing the information in multimodal data that is similar to their textual counterparts (such as captions or associated text) while neglecting the complementary information in multimodal data. For example, as shown in Figure \ref{fig:rag-example}, a textual description paired with an image might reference the {\it Rainbow Bridge}. While the image can indeed provide similar visual details of {\it Rainbow Bridge}, supplementary information such as the nightscape and the color of the bridge can also be important. For the representation-based approach, extracting and preserving such complementary information in multimodal data can enhance the quality of the responses to queries that cannot be fully resolved using textual information alone.

Inspired and motivated by the above insights, we propose a novel multimodal retrieval approach that involves {\bf C}omplementary {\bf I}nformation {\bf E}xtraction and {\bf A}lignment, denoted as {\bf \ours}. Specifically, we adopt a language model to transform text into a latent space, while employing the CLIP model~\cite{clip,li2023blip} and a projector to map image information into the same unified latent space. Then, we design a complementary information extractor, which identifies the differences in representations between the images and the text in the documents.  Based on these differences, we update the representations of the images to integrate complementary information. Furthermore, we introduce a novel optimization method tailored for \ours by constructing two complementary contrastive loss functions: one that ensures the semantics integrity of the learned representations, and another that enhances the extraction of complementary information from the images.

Extensive experiments are conducted to demonstrate the effectiveness of \ours in multimodal retrieval. The experimental results show that the proposed method achieves noticeable improvements compared to both divide-and-conquer models and universal dense retrieval models. An ablation study is carried out to highlight the contributions of the different components of \ours. Besides, we provide further discussions on the effect of the language model and include some case studies for a better understanding of \ours.

\begin{figure*}[!htbp]
    \centering
    \includegraphics[width=0.96\textwidth]{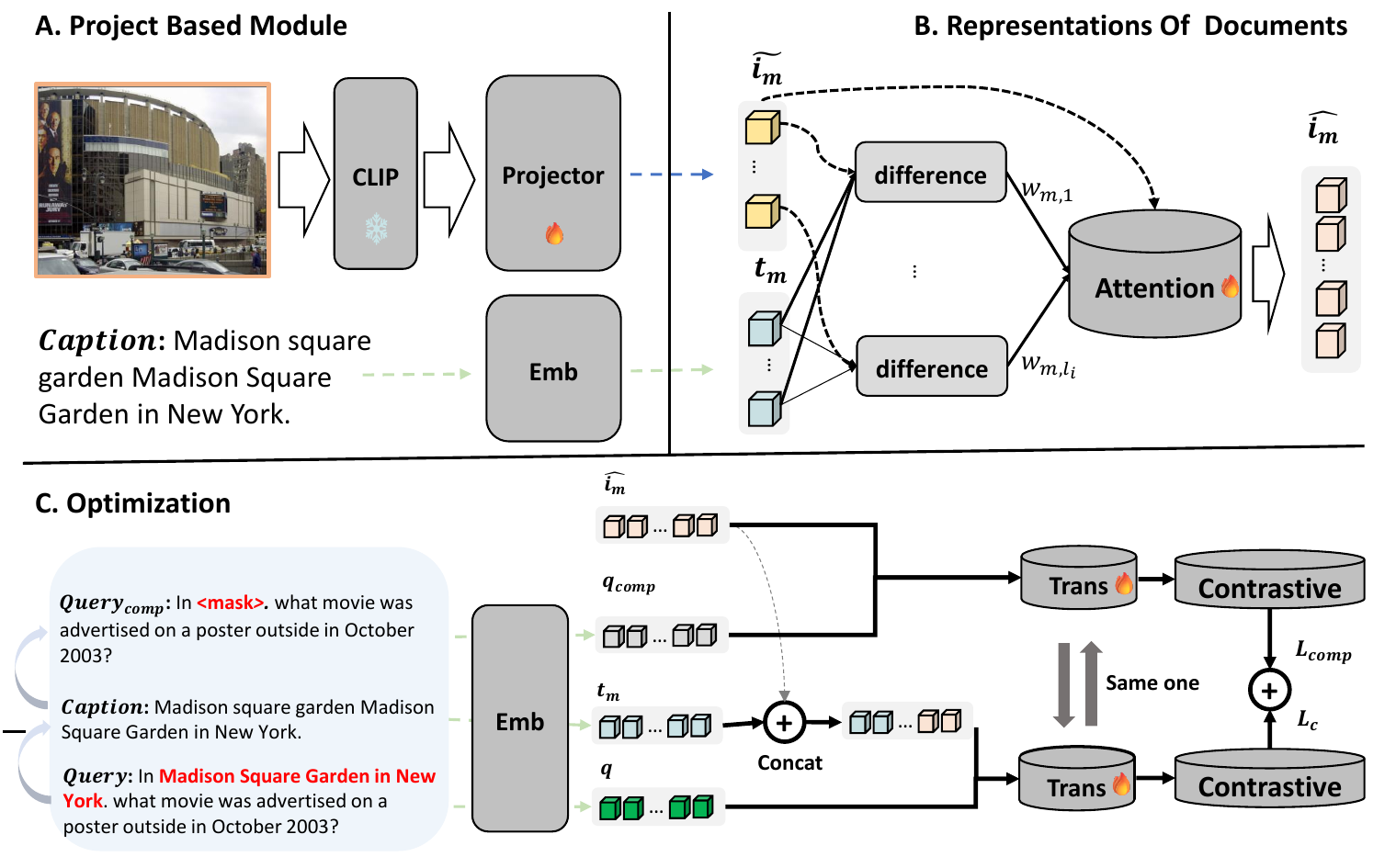}
    \caption{The overall architecture of \ours. The upper part illustrates how queries and multimodal documents are transformed into a unified latent space, while the lower part details the optimization process of \ours.}
    \label{fig:framework}
\end{figure*}

\section{Related Work}
\label{sec:related_work}
Conventional retrieval models \cite{lewis2020retrieval,t5ance} encode queries and documents into vectors via language models~\cite{chen2024benchmarking}, trained with contrastive learning and retrieved using KNN \cite{wang2022contrastive}.
Multimodal retrieval, compared to text-only retrieval, incorporates multiple modalities of information, necessitating effective utilization strategies for information from different modalities \cite{yuan2021multimodal}. One approach involves using caption models to convert information from other modalities into text, effectively converting multimodal retrieval tasks into text-only retrieval tasks \cite{baldrati2023zero}. However, such an approach may lead to information loss~\cite{che2023enhancing, li2024monkey}. 

Another approach is to employ representation models, where visual and textual encoders encode information separately before fusion \cite{clip}.
Early studies adopt a divide-and-conquer strategy, encoding each modality separately and concatenating vectors to fuse information, which potentially causes modality competition. To tackle this, UniVL-DR \cite{liu2023universal} proposes a universal multimodal retrieval framework by encoding queries and documents into a unified embedding space for retrieval, routing, and fusion. Some existing studies~\cite{wanggit,wang2021simvlm} involve training large models to unify text and visuals. However, differing representations complicate the acquisition of sufficient data for effective semantic understanding \cite{lu2024chameleon}.

Recent studies~\cite{zhou2024marvel, wang2024unified, li2024improving, li2023blip} propose a project-based framework, which leverages models like CLIP~\cite{clip} to convert visual inputs into feature sequences and introduce projector layers to align these sequences with language model embeddings. Such a framework facilitates the comparison of visual and textual content at the embedding level and converts visual information into language model ``tokens’’, capitalizing on knowledge infused during the training of language models. For example, MARVEL \cite{zhou2024marvel} utilizes the project-based framework for capturing multimodal information within the output space of the language model; MCL \cite{li2024improving} adds a retrieval token for enhancing the model's performance in representation learning.

Although remarkable progress has been made, visual information remains a novel and underexplored component for language models, which motivates us to provide effective solutions for processing visual information to attain a comprehensive representation for enhancing multimodal retrieval.

\section{Methodology}
\label{sec:methodology}

\subsection{Preliminary}
In this study, we focus on multimodal retrieval, which aims to find one or more relevant documents from a knowledge base in response to a given text query $q$. A knowledge base consisting of a total of $N$ documents can be denoted as $D = \{ d_1, d_2, \ldots, d_N \}$. Each document might contain text and images, i.e., $d_m= \{t_m, i_m\}$, where $t_m$ and $i_m$ represent the text and images in the document, respectively.

The main objective of multimodal retrieval is to align the representations of the query with the corresponding multimodal documents, ensuring that similar items are closely matched in a unified latent space. To achieve this, a multimodal encoder is employed to encode both the query and the documents, transforming them into dense representations, which can be given as:
\begin{align}
\vec{q} &= \text{Encoder}(q), \nonumber\\
\vec{d}_m &= \text{Encoder}(d_m), \forall d_m \in D.
\end{align}
After that, the similarities between the query and documents are measured via cosine similarity: 
\begin{equation}
\cos(\vec{q}, \vec{d}_m) = \frac{\vec{q} \cdot \vec{d}_m}{\|\vec{q}\| \|\vec{d}_m\|}.
\end{equation}

In the following subsections, we provide a detailed introduction of the proposed {\bf C}omplementary {\bf I}nformation {\bf E}xtraction {\bf A}lignment, denoted as {\bf \ours}. The overall architecture of \ours is illustrated in Figure~\ref{fig:framework}. 
Specifically, we first encode the queries and multimodal documents via a language model and a CLIP model, as shown in Section~\ref{section:representation}. Then, in Section~\ref{section:complementary}, we design a complementary information extractor to capture the complementary information contained in the images. The optimization designed for the proposed \ours is introduced in Section~\ref{section:optimization}.

\subsection{Representations of Queries and Multimodal Documents} 
\label{section:representation}
Firstly, as shown in part A of Figure~\ref{fig:framework}, we transform both the queries and the multimodal documents into their dense representations. Inspired by previous studies~\cite{zhou2024marvel,li2024improving,li2023blip}, we utilize a transformer-based language model as the backbone of the encoder to encode both text and images. A transformer-based language model typically consists of an embedding layer followed by several transformer blocks. We denote the embedding layer as $\text{Emb}(\cdot)$ and the transformer blocks as $\text{Trans}(\cdot)$.

The text query $q$ can be transformed into dense representations via language models, which can be given as:
\begin{equation}
\vec{q} = \text{Trans}(\text{Emb}(q)),\label{eq:q_emb}
\end{equation}
where $\vec{q} \in \mathbb{R}^{l_q \times d}$, $l_q$ denotes the number of tokens in the query, and $d$ denotes the dimension of the dense representations.

For a document containing both text and images, i.e., $d_m = \{t_m, i_m\}$, the text $t_m$ and image $i_k$ are processed separately before being fused. Specifically, we begin by feeding the text within the document to the embedding layer:
\begin{equation}
    \vec{t}_m = \text{Emb}(t_m),
\end{equation}
where $\vec{t}_m = [\vec{t}_m^{(1)}, \vec{t}_m^{(2)},\ldots , \vec{t}_m^{(l_t)}] \in \mathbb{R}^{l_t \times d}$ and $l_t$ denotes the number of token in the text. 

The images in the documents can typically be represented as RGB multi-channel matrices. We adopt a frozen CLIP~\cite{clip} visual encoder to transform these matrices into a set of semantic vectors. Such a process involves dividing an image into multiple patches, with each patch representing a different region of the image, which can be formally given as:
\begin{equation}
\vec{i}_m = \text{CLIP}_\text{Visual}(i_m),
\end{equation}
where $\vec{i}_m \in \mathbb{R}^{l_i \times d_{\text{clip}}}$, $l_i$ denotes the number of patches, and $d_{\text{clip}}$ denotes the hidden dimension of CLIP visual encoder.

Note that the hidden dimension of these image representations might be different from that of text representations, i.e., $d_{\text{clip}} \neq d$. Therefore we adopt a linear layer as the projector for further alignment:
\begin{equation}
\widetilde{\vec{i}_m} = \text{Proj}(\vec{i}_m), \label{eq:projector}
\end{equation}
where $\widetilde{\vec{i}_m} = [\widetilde{\vec{i}_m}^{(1)}, \widetilde{\vec{i}_m}^{(2)},\ldots , \widetilde{\vec{i}_m}^{(l_i)}] \in \mathbb{R}^{l_i \times d}$.

\subsection{Complementary Information Extractor}
\label{section:complementary}
The proposed \ours aims to capture the complementary information contained in images, i.e., the information not encompassed by the text in a document. To achieve this, we design a complementary information extractor that captures the differences between images and textual content. Specifically, we calculate the patch-level distances between text and image representations, identifying the maximum value to serve as a difference measurement, which is similar to those adopted in BLIP-2~\cite{li2023blip}. Formally, the difference between the $j$-th patch of image and the text within document $m$, denoted as $r_{m,j}$, can be calculated as:
\begin{equation}
r_{m,j} = -\max_{c \in [l_t]}\ \text{cos}(\widetilde{\vec{i}_m}^{(j)}, \vec{t}_m^{(c)}).
\end{equation}
The obtained differences are normalized into a range of $[0, 1]$ for effectively weighing the image patches, as given by:
\begin{equation}
w_{m,j} = \frac{1 + r_{m,j}}{2}.
\end{equation}

We adopt an attention layer for re-weighting the patches of the image based on the calculated $\vec{w}_m = w_{m,j}\ \forall j \in [l_i]$. The attention scores of embeddings are multiplied by the corresponding weights for normalization, which can be given as:
\begin{equation}
\vec{Q}_m=\widetilde{\vec{i}_m}\vec{W}_Q,\ \vec{K}_m=\widetilde{\vec{i}_m}\vec{W}_K,\ \vec{V}_m=\widetilde{\vec{i}_m}\vec{W}_V,\nonumber
\end{equation}
\begin{equation}
\label{eq:attention}
\widehat{\vec{i}_m}=\text{softmax}(\frac{\vec{Q}_m \vec{K}_m \cdot \vec{w}_m}{\sqrt{d}})\vec{V}_m,
\end{equation}
where $ \vec{W}_Q, \vec{W}_K, \vec{W}_V$ are learnable matrices and ``$\cdot$'' here stands for the broadcasting method.

After that, as shown in part B of Figure~\ref{fig:framework}, the embeddings of image and text are concatenated together and fed into transformer blocks for multimodal knowledge fusion. The obtained dense representation of the document $d_m$ can be given as:
\begin{equation}
\vec{d}_m = \text{Trans}([\vec{e_\text{start}}\oplus\widehat{\vec{i}_m}\oplus\vec{e_\text{end}}\oplus\vec{t}_m]),
\label{eq:trans}
\end{equation}
where $\oplus$ denotes the concatenate operation, and $\vec{e_\text{start}}$ and $\vec{e_\text{end}}$ denote the embeddings of special tokens {\it <start>} and {\it <end>}, respectively, for distinguishing the text and image embeddings explicitly.

Finally, for a given query, we rank the documents based on the similarities between their representations (refer to Eq.\eqref{eq:trans}) and the query representation (refer to Eq.\eqref{eq:q_emb}) to obtain the relevant documents.

\begin{table*}[t]
    \centering
    \caption{Comparisons on the WebQA-Multi dataset. The best results are marked with bold.}
    \resizebox{\linewidth}{!}{
    \begin{tabular}{lcccccc}
        \toprule
          & \textbf{mrr@10} & \textbf{ndcg@10} & \textbf{mrr@20} & \textbf{ndcg@20} & \textbf{rec@20} & \textbf{rec@100} \\ 
        \midrule
         BM25                & 22.11  & 22.92  & 22.80  & 25.41  & 46.27  & 62.82  \\
        CLIP-DPR           &37.35 & 37.56 & 37.93  & 40.77  & 69.38  & 85.53  \\
     BM25 \& CLIP-DPR  & 42.27& 41.58  & 42.79 & 44.69  & 73.34 & 87.50  \\
        BM25 \& CLIP (oracle) & 61.05 & 58.18  & 61.37 & 60.45  & 80.82 & 90.83 \\
        \midrule
        VinVL-DPR         & 38.14  & 35.43  & 38.74  & 37.79  & 53.89  & 69.42  \\
     CLIP-DPR          & 48.83  & 46.32  & 49.34  & 49.11  & 69.84  & 86.43  \\
        UniVL-DR          & 62.40 & 59.32 & 62.96 & 61.22  &80.37 & 89.42  \\
        T5-ANCE & 64.13 & 62.03 & 64.40 & 64.02 & 83.81 & 92.07 \\
     MARVEL             & 65.43& 63.07 & 65.67 & 65.00  & 84.32& 92.27  \\
     \midrule
         \textbf{CIEA (ours)}             & \textbf{66.16} & \textbf{63.89} & \textbf{66.41} & \textbf{65.85} & \textbf{85.43} & \textbf{92.75} \\ 
        \bottomrule
        
    \end{tabular}
    }
    \label{webqa-multi}
\end{table*}
\subsection{Optimization} 
\label{section:optimization}
The trainable parameters of \ours framework include the projector (as defined in Eq.~\eqref{eq:projector}), the added attention weights $ \vec{W}_Q, \vec{W}_K, \vec{W}_V $ for re-weighting (as defined in Eq.~\eqref{eq:attention}), and the transformer layers of the language models (as defined in Eq.~\eqref{eq:trans}). In this section, we introduce how to optimize these trainable parameters in \ours.

Firstly, we identify the text segments from the queries in the training set that correspond to the text in the ground truth documents. For example, as shown in Figure~\ref{fig:framework}, we mask out ``{\it Madison Square Garden in New York}'' that appears in the caption. We match the token IDs generated by the tokenizer of the language model for the query and the text in document for such identification.

After that, we replace these identified text segments with special tokens {\it <mask>} while preserving the remaining parts of the query. We denote the processed query as $q_\text{comp}$, whose representations can be obtained in a similar manner to the computation defined in Eq.~\eqref{eq:q_emb}:
\begin{equation}
\vec{q}_{\text{comp}} = \text{Trans}(\text{Emb}(q_{\text{comp}})).\label{eq:image_query}
\end{equation}

We also construct negative samples for the queries to apply a contrastive loss for optimization. The documents annotated as relevant to the query are denoted as $D^+$, while $N$ negative documents are sampled from the knowledge base and denoted as $D^- = \{d_1^-, d_2^-, \ldots, d_{N}^-\}$.
The contrastive loss can be defined as:
\begin{equation}
\mathcal{L}_c = -\log \frac{e^{\cos(\vec{q}, \vec{d^+}) / \tau}}{e^{\cos(\vec{q}, \vec{d^+}) / \tau} + \sum_{D^-} e^{\cos(\vec{q}, \vec{d^-}) / \tau}},
\end{equation}
where $\tau$ denotes the temperature, $\cos(\cdot)$ denotes the cosine similarity function, $\vec{q}$ denotes the representation of the query, $\vec{d^{+}}$ and $\vec{d^{-}}$ denote the representations of positive documents and negative documents, respectively.

Meanwhile, to enhance the extraction of complementary information from images within the documents, we construct another contrastive loss function based on $\vec{q}_{\text{comp}}$ and the representations of images, which can be given as:
\begin{equation}
\mathcal{L}_\text{comp} =  -\log  \frac{e^{\cos(\vec{q}_{\text{comp}}, \vec{d}_{\text{img}}^+) / \tau}}{e^{\cos(\vec{q}_{\text{comp}}, \vec{d}_{\text{img}}^+) / \tau} + \sum_{D^-} e^{\cos(\vec{q}_{\text{comp}}, \vec{d}_{\text{img}}^-) / \tau}},\label{eq:l_comp}
\end{equation}
where $\vec{d}_\text{img} = \text{Trans}(\widehat{\vec{i}_k})$ denotes the representations of the images in the documents.
The intuition behind the above loss function is that we promote the representations of images to align with the information contained in the query that is not present in the text within documents.

Finally, the training objective of the proposed \ours can be given as:
\begin{equation}
\mathcal{L} = \mathcal{L}_c + \lambda \cdot \mathcal{L}_{\text{comp}},
\end{equation}
where $\lambda$ is a hyperparameter that balances the contributions of the two contrastive losses.

\section{Experiments}
\label{sec:exp}

\subsection{Settings}
\label{subset:exp_settings}

\begin{table*}[t]
    \centering
    \caption{Comparisons on the WebQa-Image and EDIS. The best results are marked with bold.}
    \resizebox{\linewidth}{!}{
    \begin{tabular}{lcccccccc}
        \toprule
        & \multicolumn{4}{c}{\textbf{WebQa-Image}} & \multicolumn{4}{c}{\textbf{EDIS}} \\
        \cmidrule(lr){2-5} \cmidrule(lr){6-9}
        & \textbf{mrr@10} & \textbf{ndcg@10} & \textbf{mrr@20} & \textbf{ndcg@20} 
        & \textbf{mrr@10} & \textbf{ndcg@10} & \textbf{mrr@20} & \textbf{ndcg@20} \\ 
        \midrule
        CLIP-DPR & 59.78 & 61.05 & 60.2 & 63.28 & 62.52 & 39.19 & 62.90 & 44.24 \\
        UniVL-DR & 65.95 & 67.33 & 66.24 & 69.01 & 62.89 & 39.50 & 63.29 & 44.54 \\
        T5-ANCE & 64.38 & 65.65& 64.73 & 67.52 & 63.95 & 39.74 & 64.30 & 43.76 \\
        MARVEL-DPR & 62.51 & 63.48 & 62.90 & 65.54 & 63.60 & 37.76 & 63.98 & 41.89 \\
        MARVEL & 66.43 & 67.60 & 66.76 & 69.40 &  67.00& 42.19 & 67.28 & 46.63 \\
        \midrule
        
        \textbf{CIEA (ours)} & \textbf{67.40} & \textbf{68.77} & \textbf{67.74} & \textbf{70.51} & \textbf{68.11} & \textbf{42.57} & \textbf{68.42} & \textbf{47.19}  \\
        \bottomrule
    \end{tabular}
    }
    \label{webqa-img}
    
\end{table*}

\paragraph{Datasets}
We conduct extensive experiments on the WebQA ~\cite{chang2022webqa} and EDIS~\cite{liu2023edis} dataset to compare the proposed method against baselines. WebQA is an open-domain multimodal question answering dataset in which each query is associated with one or more relevant documents. There are 787,697 documents containing only textual information, while 389,750 documents contain both text and images. For clarity, we refer to this dataset as {\bf WebQA-Multi} in this study. Besides, to better show the effectiveness of multimodal retrieval, we extract all documents containing images and construct the {\bf WebQA-Image} dataset. {\bf EDIS} retrieves images via textual queries, with 26,000 training, 3,200 development, and 3,200 test samples from 1M image corpus (each paired with a text caption). More statistics of datasets can be found in Appendix~\ref{appendix:dataset}.

\paragraph{Metrics}
To evaluate the effectiveness of the proposed method and baselines in multimodal retrieval, we adopt six widely-used metrics, including MRR@10, MRR@20, NDCG@10, NDCG@20, REC@20, and REC@100. The MRR~\cite{bajaj2016ms} and NDCG~\cite{taylor2008softrank} metrics are calculated using the official code\footnote{https://github.com/microsoft/MSMARCO-Passage-Ranking/blob/master/ms\_marco\_eval.py}.

\paragraph{Implementation Details}
In the experiments, we employ T5-ANCE~\cite{t5ance} as the foundational language model and CLIP as the visual understanding module to implement the proposed CIEA model to enhance the image understanding capabilities. The document representation capabilities of T5-ANCE are enhanced by using ANCE~\cite{xiongapproximate} to sample negative documents based on T5~\cite{raffel2020exploring}. We truncate the text input length to 128 tokens. During training, we adopt the AdamW~\cite{loshchilov2017fixing} optimizer with a maximum of 40 epochs, a batch size of 64, a learning rate of 5e-6, and set the temperature hyperparameter $\tau$ to 0.01. The hyperparameter $\lambda$ is set to 0.0011 for WebQA-Multi, 0.019 for WebQA-Image, and 0.001 for EDIS, respectively. More implementation details can be found in Appendix~\ref{appendix:imple}.

\subsection{Baselines}\label{subsec:baselines}
We compare CIEA with two different types of baselines: divide-and-conquer models and universal dense retrieval models.

{\bf Divide-and-conquer models} suggest retrieving image and text documents separately and then merging the retrieval results. We employ various single-modal retrievers to instantiate the divide-and-conquer models, including VinVL-DPR~\cite{zhang2021vinvl}, CLIP-DPR~\cite{clip}, and BM25~\cite{bm25}. The results of multi-modality retrieval are combined based on their uni-modal rank reciprocals or oracle modality routing. The latter approach demonstrates the maximum potential performance of our divide-and-conquer models in retrieval tasks.

Regarding {\bf universal dense retrieval models}, we adopt the following recent studies as baselines: (i) Pre-trained multimodal alignment models VinVL-DPR~\cite{zhang2021vinvl} and CLIP-DPR~\cite{clip}, which are trained using the DPR framework; (ii) UniVL-DR~\cite{liu2023universal}, which employs modality-balanced hard negatives to train text and image encoders, and uses image language adaptation techniques to bridge the modality gap between images and text; (iii) MARVEL~\cite{zhou2024marvel}, a project-based approach that encodes image documents with CLIP and projects them into the embedding layer of a language model for fusion.

\begin{table*}[!t]
    \centering
    \caption{Experimental results when using different language models. The {\it Text only} setting denotes we only use the document's captions for training without incorporating image information, while the {\it Project-based} setting is the most common projection method without visual alignment.}
    \resizebox{\linewidth}{!}{
    \begin{tabular}{lcccccccc}
        \toprule
        \textbf{Language model} & \textbf{setting} & \textbf{mrr@10} & \textbf{ndcg@10} & \textbf{mrr@20} & \textbf{ndcg@20} & \textbf{rec@20} & \textbf{rec@100} \\ \midrule
        \multirow{3}{*}{T5-ANCE} & Text only & 64.13 & 62.03 & 64.40 & 64.02 & 83.81 & 92.07 \\ 
        & Project-based & 65.43 & 63.07 & 65.67 & 65.00 & 84.32 & 92.27 \\ 
        & \textbf{CIEA} & \textbf{66.16} & \textbf{63.89} & \textbf{66.41} & \textbf{65.85} & \textbf{85.43} & \textbf{92.75}  \\ \midrule

        \multirow{3}{*}{BERT} & Text only & 61.39 & 58.20 & 61.69 & 60.32 & 79.59 & 89.10 \\ 
        & Project-based & 63.03 & 60.20 & 63.31 & 62.37 & \textbf{81.80} & \textbf{90.78} \\ 
        & \textbf{CIEA} & \textbf{63.63} & \textbf{60.46} & \textbf{63.90} & \textbf{62.49} & 81.51 & 90.57 \\ \midrule

        \multirow{3}{*}{BART} & Text only & 60.14 & 56.82 & 60.49 & 59.14 & 78.41 & 89.25 \\ 
        & Project-based & 62.67 & 59.70 & 62.98 & 62.04 & 81.77 & 90.77 \\ 
        & \textbf{CIEA} & \textbf{63.73} & \textbf{60.43} & \textbf{64.04} & \textbf{62.72} & \textbf{81.96} & \textbf{90.98} \\ \midrule

        \multirow{3}{*}{GPT2} & Text only & 54.68 & 51.49 & 55.07 & 53.77 & 72.91 & 84.86 \\ 
        & Project-based & 58.59 & 55.00 & 58.92 & 57.15 & 75.84 & 86.79 \\ 
        & \textbf{CIEA} & \textbf{59.25} & \textbf{55.35} & \textbf{59.59} & \textbf{57.62} & \textbf{76.47} & \textbf{86.92} \\ \midrule

        \multirow{3}{*}{GPT2-LARGE} & Text only & 63.58 & 60.44 & 63.86 & 62.45 & 81.24 & 89.98 \\ 
        & Project-based & 64.33 & 61.98 & 64.60 & 62.04 & 83.97 & 92.11 \\ 
        & \textbf{CIEA} & \textbf{65.38} & \textbf{63.00} & \textbf{65.62} & \textbf{64.96} & \textbf{84.96} & \textbf{92.68} \\ \bottomrule
    \end{tabular}
    }
    \label{tab:language_models}
\end{table*}

\begin{table}[!t]
    \caption{Ablation study on WebQa-multi. Here, "w/o" stands for "without," and "Base" refers to the setting where both modules are removed.}
    \centering
    \resizebox{\linewidth}{!}{
    \begin{tabular}{lcccc}
        \toprule
          & \textbf{mrr@10} & \textbf{ndcg@10} & \textbf{rec@20} & \textbf{rec@100} \\ 
        \midrule
         \textbf{\ours}& \textbf{66.03} & \textbf{63.70} & \textbf{85.14} & \textbf{92.63} \\ 
         
         w/o image query           & 65.66 & 63.44 & 84.74 & 92.54 \\ 
         w/o attention           & 65.90 & 63.39 & 84.95 & 92.60 \\ 
         Base & 65.30& 63.19 & 84.49& 92.40 \\
         
        \bottomrule
    \end{tabular}
    }
    \label{table:aba}
\end{table}

\paragraph{Comparisons}
The experimental results on WebQA and EDIS are shown in Table \ref{webqa-multi} and \ref{webqa-img}, respectively. From these results, we can observe that the proposed CIEA achieves noticeable improvements across various metrics compared to both divide-and-conquer models and universal dense retrieval models. CIEA achieves outperformance on both WebQA and EDIS, demonstrating its effectiveness in capturing crucial information from both text and images for enhancing multimodal retrieval. As shown in Tables \ref{webqa-multi}, compared to universal dense retrieval models that employ projectors for language model grounding (e.g., MARVEL, which shows improvement over text-only methods with a 1.3 MRR@10 improvement over T5-ANCE), \ours achieves a further 0.74 MRR@10 improvement, with analogous enhancements replicated in Table \ref{webqa-img}. These CIEA-driven advancements underscore the critical benefits of complementary information alignment - a framework that enables superior visual information modeling and enhances the representational power of project-based models.

\paragraph{Ablation Study}
We conduct an ablation study to demonstrate the contributions of the image query and the attention layer.  Specifically, we remove the attention layer designed for re-weighting (refer to Eq.~\eqref{eq:attention}) and the utilization of image query (refer to Eq.~\eqref{eq:l_comp}), respectively. We also perform {\it base} setting by removing these two components. 

The experimental results are shown in Table~\ref{table:aba}. 
Although the removal of the image query does not lead to a decrease in rec@100, other metrics exhibit varying degrees of decline, particularly rec@20, which drops from 85.14 to 84.74. On the other hand, the removal of the attention mechanism results in larger significant decreases in  ndcg@10 metrics, highlighting the importance of complementary information extraction. Further, we also compare the baseline configuration with the simultaneous removal of both modules. The results indicate that removing a single module yields improvements in various metrics compared to the baseline, demonstrating that both the image query and the attention layer contribute to enhancing retrieval accuracy in multimodal retrieval. More results can be found in Appendix \ref{appendix:statistical}.

\begin{table*}[!th]
    \centering
    \caption{Experimental results with different weights on WebQA-Multi.}
    \begin{tabular}{lcccc}
    \hline
    Weights        & MRR@10   & NDCG@10  & REC@20   & REC@100  \\
    \hline
    Dissimilar        & \textbf{66.03±0.012} & \textbf{63.70 ± 0.030} & \textbf{85.14±0.052} & \textbf{92.63±0.008} \\
    Similar  & 65.60±0.038 & 63.35±0.005 & 85.00±0.012 & 92.61±0.003 \\
    \hline
    \end{tabular}
    \label{atten_weight}
\end{table*}

\subsection{Further Discussions}
\label{subsec:discussion}
\paragraph{The Effects of Language Models in \ours}
To provide further discussions regarding the effects of language models used in the proposed CIEA, we conduct experiments with different language models as the backbone, including T5-ANCE~\cite{t5ance}, BART~\cite{lewis2019bart}, BERT~\cite{lee2018pre}, GPT-2, and GPT-2-LARGE ~\cite{radford2019language}. These language models cover three mainstream architectures, i.e., encoder-decoder, encoder-only, and decoder-only. More implementation details can be found in Appendix~\ref{appendix:language}.

The experimental results in Table~\ref{tab:language_models} demonstrate the significant impacts of the backbone language model on multimodal retrieval effectiveness. For text-only retrieval, language models with larger parameter sizes, such as GPT-2 LARGE and T5-ANCE, achieve better overall performance. Both project-based models and CIEA benefit from powerful backbones, showing performance gains through image-derived information. Notably, CIEA maintains competitive results across different backbones compared to project-based approaches, confirming its robustness.

\paragraph{Attention Weight Selection}

The application of maximum cosine similarity for computing patch-level similarities is designed to extract image regions exhibiting lower correspondence with textual descriptions. To determine the optimal extraction approach, we perform empirical experiments that retain original cosine values rather than their complements. As demonstrated in Table \ref{atten_weight}, explicit dissimilarity computation has proven more effective for capturing fine-grained patch-level image-text mappings.

\paragraph{Computational Efficiency}
Although the proposed dual losses might increase training complexity, the computation of the losses is relatively independent and would not lead to a multiplicative increase in overall complexity. The proposed method achieves similar computational efficiency compared to MARVEL, which is also the project-based method. For example, with the sample devices, the training on the WebQA-Multid dataset with 4,966 samples with MARVEL needs around 6.1 minutes while that of \ours is around 6.5 minutes.

\begin{figure*}[!t]
    \centering
    \includegraphics[width=\linewidth]{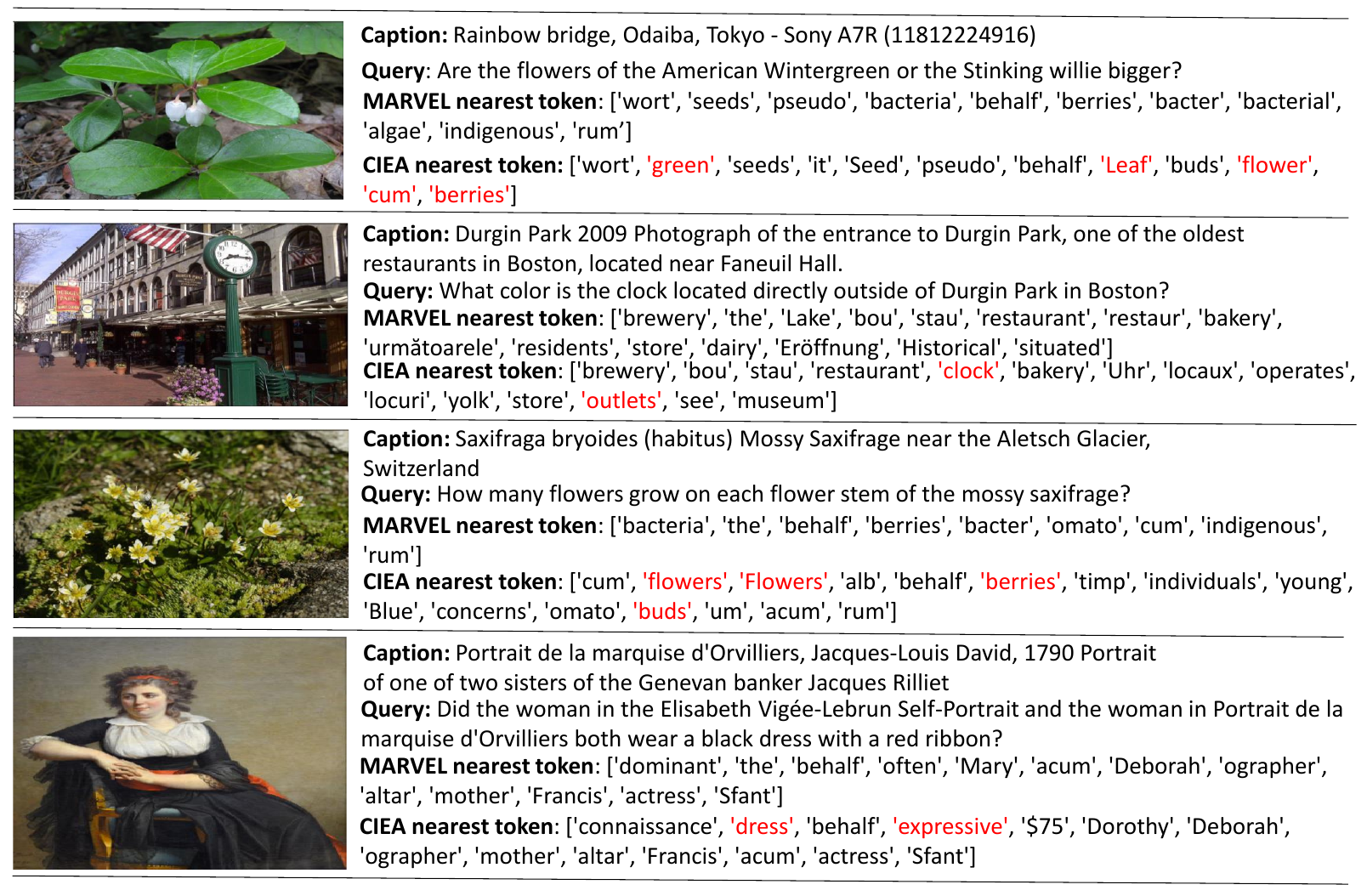}
    \caption{Case studies. The nearest token represents the embeddings in the vocabulary that are closest to the visual embeddings, with duplicates removed. The words in red represent terms related to the image that are not found in MARVEL within CIEA.}
    \label{fig:case}
\end{figure*} 

\paragraph{Case Study}
We conduct case studies on the WebQA dataset for a better understanding. We apply cosine similarity to identify the words from the language model's vocabulary that are closest to the embeddings projected from images, showing what information is extracted from images. We compare the proposed CIEA and the strongest baseline MARVEL.
The results are illustrated in Figure~\ref{fig:case}, which indicates that MARVEL focuses on image information that is close to the text, while CIEA captures more supplemental information from images. For example, CIEA captures some terms, such as {\it green}, {\it leaf}, and {\it clock}, that are semantically related to the image but are not mentioned in the caption. The term {\it clock} in the second case and {\it flowers} in the third case are relevant to the query, which can assist the model in efficiently locating and retrieving relevant content. These results further confirm the advances of CIEA for multimodal retrieval. 
Discussions regarding the failure cases are provided in Appendix \ref{appendix:failure}, which further illustrate the model's behavior in challenging scenarios and inspire future improvements.

\section{Conclusion}
In this paper, we propose \ours to enhance the effectiveness of multimodal retrieval. The main idea of \ours is to enhance the capturing of complementary information in multimodal data. Specifically, \ours utilizes language models and CLIP to transform multimodal documents into a unified latent space. For complementary information extraction, we calculate the patch-level distances between text and images, which are then used to re-weight the image representations. Regarding the optimization of \ours, besides applying a contrastive loss for learning the semantics of text, we also encourage the alignment of image representations with the complementary information in queries. We conduct a series of experiments to demonstrate the advantages of \ours compared to two different types of multimodal retrieval approaches. Further discussions on the effect of language models show the robustness of \ours, and several case studies are included for better understanding.

\section*{Limitations}
Multimodal retrieval is currently in a phase of rapid development. In this paper, our exploration is limited to queries that contain only text, with a focus on extracting information from images. Expanding the proposed approach to include more types of multimodal data, such as audio and video, is a promising direction for future research. Besides, although we conduct experiments with various language models, we are constrained by computational resources and have not yet explored larger models. 

\section*{Acknowledgement}
This work is supported by Key-Area Research and Development Program of Guangdong Province (No.2024B1111100001).
\bibliography{custom}

\appendix
\newpage
\section{Datasets }\label{appendix:dataset}
\begin{table}[ht]
    \centering
    \caption{Statistics of the adopted WebQA.}
    \resizebox{\linewidth}{!}{
    \begin{tabular}{ccccc}
        \toprule
        \multirow{2}{*}{Modality} & \multirow{2}{*}{Documents} & \multicolumn{3}{c}{Queries}  \\ 
        & & Train     & Dev       & Test         \\ \midrule
        Image    & 389,750     & 16,400    & 2,554     & 2,511     \\
        Text     & 787,697     & 15,366    & 2,446     & 2,455     \\ 
        Total    & 1,177,447   & 31,766    & 5,000     & 4,966     \\ \bottomrule
    \end{tabular}
    }
    \label{apptable:dataset}
\end{table}
WebQA is an open-domain question answering dataset, where each query is associated with one or more multimodal documents to assist in generating responses. WebQA can be divided into three partitions: train, dev, and test, with data statistics as shown in Table \ref{apptable:dataset}. The retrieval corpus contains 389,750 image documents with visual information and captions, as well as 787,697 plain text documents. Our partitioned WebQA-Image focuses solely on image documents for testing the effectiveness of \ours in multimodal modeling. In contrast, WebQA-Multi retrieves from a total of 1,177,447 documents to assess whether the model can maintain an accurate representation of text information. 

WebQA is used by many baselines, such as MARVEL~\cite{zhou2024marvel}. ClueWeb22-MM is also utilized in MARVEL. We have applied for this dataset from Carnegie Mellon University; unfortunately, our request was denied due to export control restrictions. Therefore, we also use EDIS to evaluate our methods. EDIS~\cite{liu2023edis}  is a comprehensive dataset consisting of 1 million image-text pairs sourced from Google. The dataset includes a training set of 26,000 pairs, accompanied by validation and test sets, each containing 3,200 pairs. With a high entity count of 4.03, EDIS reflects a diverse range of semantic content, making it a valuable resource for research in image-text retrieval tasks.

\begin{table*}[!t]
    \centering
    \caption{Results of different language models on DPR training framework. }
    \resizebox{\linewidth}{!}{
    \begin{tabular}{lcccccccc}
        \toprule
        \textbf{Language model} & \textbf{setting} & \textbf{mrr@10} & \textbf{ndcg@10} & \textbf{mrr@20} & \textbf{ndcg@20} & \textbf{rec@20} & \textbf{rec@100} \\ \hline
        \multirow{3}{*}{T5-ANCE} & Text only & 52.67 & 45.69 & 53.08 & 48.48 & 67.70 & 82.07 \\
        & Project-based & 56.91 & 54.26 & 57.31 & 56.88 & 77.20 & 89.26 \\ 
        & \textbf{CIEA} & \textbf{57.49} & \textbf{54.80} & \textbf{57.86} & \textbf{57.24} & \textbf{77.21} & \textbf{89.41}  \\ \hline

        \multirow{3}{*}{BERT} & Text only & 39.50 & 36.23 & 40.06 & 38.69 & 55.37 &72.72\\ 
        & Project-based & 52.65 & 49.41 & 53.21 & 51.97 & 71.16 & 85.40 \\ 
        & \textbf{CIEA} & \textbf{53.40} & \textbf{49.96} & \textbf{53.83} & \textbf{62.56} & \textbf{71.95} & \textbf{85.72} \\ \hline

        \multirow{3}{*}{BART} & Text only & 24.52 & 21.79 & 25.04 & 23.98 & 36.91 & 54.66 \\ 
        & Project-based & 50.46 & 47.72 & 50.94 & 50.04 & 69.40 & 83.69 \\ 
        & \textbf{CIEA} & \textbf{50.97} & \textbf{48.06} & \textbf{51.43} & \textbf{50.72} & \textbf{70.44} & \textbf{84.81} \\ \hline

        \multirow{3}{*}{GPT2} & Text only & 20.38 & 18.62 & 20.86 & 20.19 & 29.97 & 44.33 \\ 
        & Project-based & \textbf{45.94} & 42.45 & 46.46 & 45.08 & 63.16 & 78.34 \\ 
        & \textbf{CIEA} & 45.82 & \textbf{42.60}& \textbf{46.51} & \textbf{45.13}  & \textbf{63.46} & \textbf{78.91} \\ \hline

        \multirow{3}{*}{GPT2-LARGE} & Text only & 37.25 & 34.52 & 37.67 & 37.11 & 54.37 & 73.08 \\ 
        & Project-based & 51.44 & 48.98 & 51.89 & 51.64 & 72.06 & 86.61 \\
        & \textbf{CIEA} & \textbf{54.28} & \textbf{51.81} & \textbf{54.68} & \textbf{54.39} & \textbf{74.94} & \textbf{88.52} \\ \hline
    \end{tabular}
    }
    \label{appendixtab:language_models}
\end{table*}
\section{Implementation Details}\label{appendix:imple}
In our experiment, we use T5-ANCE \cite{t5ance} as the base language model and utilize CLIP as the visual understanding module to implement our CIEA model, enhancing the image understanding capability of T5-ANCE. For easier comparison, we initialize our projector using the pre-trained projector from MARVEL. The visual encoder is initialized using the clip-vit-base-patch32 checkpoint from OpenAI\footnote{https://huggingface.co/sentence-transformers/clip-ViT-B-32}. We truncate the text input length to 128 tokens.

During the training process, we use the AdamW~\cite{loshchilov2017fixing} optimizer, set the maximum training epochs to 40, with a batch size of 64, a learning rate of 5e-6, and the temperature hyperparameter $\tau$ set to 0.01. We follow the setup of UniVL-DR \cite{loshchilov2017fixing} by training with the ANCE sampling method. Starting from the \ours-DPR model, fine-tuned with negative examples from within the batch, we continue to train \ours-DPR to achieve a balanced modality with difficult negative examples. For negative sampling in evaluation, we shuffle the training set and select other samples within the same batch as negative examples in DPR, while in the ANCE approach, we utilize the top 100 samples with the highest similarity retrieval by \ours-DPR as hard negatives. All models are evaluated every 500 steps, with early stopping set at 5 steps. For the parameter $\lambda$, we perform a grid search and select the parameter that yielded the lowest loss on the validation set, setting it to 0.0011 for  WebQA-Multi, 0.019 for WebQA-Image, and 0.001 for EDIS. The results of various baselines for the WebQA-Multi dataset are provided in MARVEL~\cite{zhou2024marvel}, while the experimental results for other baselines are reproduced using the open-source code from the original papers. All experiments are conducted on a single NVIDIA A100.

\section{Statistical significance}\label{appendix:statistical}
\begin{table}[h]
    \centering
    \caption{ Standard Deviation Values on WebQa-multi. }
    \resizebox{\linewidth}{!}{
    \begin{tabular}{l@{\hspace{0.5em}}c@{\hspace{0.5em}}cc@{\hspace{0.7em}}c}
    \hline
    Model        & MRR@10   & NDCG@10   & REC@100  \\
    \hline
    \textbf{\ours}        & \textbf{66.03±0.012} & \textbf{63.70±0.030} & \textbf{92.63±0.008} \\
    w/o image query & 65.66±0.008 & 63.44±0.072 &  92.54±0.003 \\
    w/o attention & 65.90±0.098 & 63.39±0.003 &  92.60±0.001 \\
    
    MARVEL        & 65.30±0.017 & 63.19±0.005  & 92.40±0.012 \\
    \hline
    \end{tabular}
    }
    \label{appendix:deviation}
\end{table}
To avoid potential concerns that the results are justified, we repeat the experiments with different random seeds and report the average values and standard deviation values in Table ~\ref{appendix:deviation}. The results demonstrate the proposed \ours achieves significant and consistent improvements compared to MARVEL. Besides, different modules make positive contributions to the overall performance of ICEA.
\begin{table*}[htbp]
\centering

\begin{tabularx}{\textwidth}{@{}c|ccc@{}}
\toprule
\multicolumn{4}{l}{\textbf{Question:} Which restaurant on Orchard Road in Singapore has a tiered orange ceiling?} \\
\midrule
\multicolumn{1}{c}{\textbf{Golden}} & \multicolumn{3}{c}{\textbf{Retrieval (Top 3)}} \\
\midrule

\raisebox{-0.5\height}{\includegraphics[width=0.23\textwidth,height=0.23\textwidth]{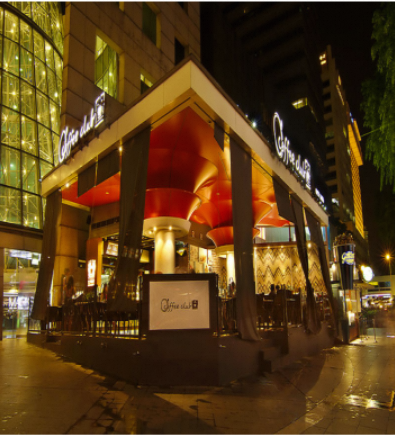}} &
\raisebox{-0.5\height}{\includegraphics[width=0.23\textwidth,height=0.23\textwidth]{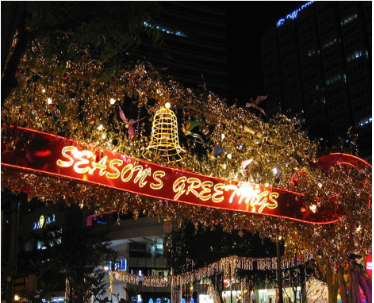}} &
\raisebox{-0.5\height}{\includegraphics[width=0.23\textwidth,height=0.23\textwidth]{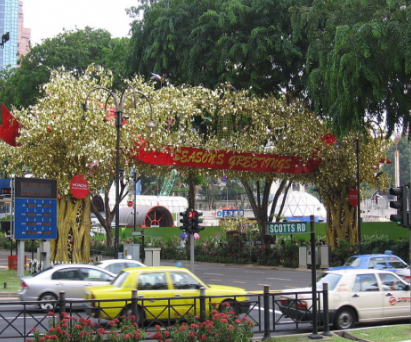}} &
\raisebox{-0.5\height}{\includegraphics[width=0.23\textwidth,height=0.23\textwidth]{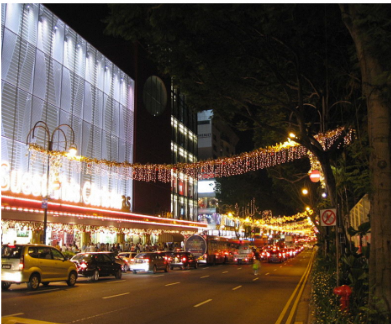}} \\

\addlinespace[1ex]  
\parbox{0.23\textwidth}{\centering Coffee Club, Orchard Road (8171776629) Coffee Club at Wheelock Place, Singapore.} &
\parbox{0.23\textwidth}{\centering Orchard Road 12, Xmas, Dec 06 Orchard Road, Christmas Light-up 2006, Singapore.} &
\parbox{0.23\textwidth}{\centering Orchard Road, Xmas, Dec 06 Orchard Road, Christmas Light-up 2006, Singapore.} &
\parbox{0.23\textwidth}{\centering Orchard Road 15, Xmas, Dec 06 Orchard Road, Christmas Light-up 2006, Singapore.} \\
\bottomrule
\end{tabularx}

\begin{tabularx}{\textwidth}{@{}c|ccc@{}}
\toprule
\multicolumn{4}{l}{\textbf{Question:} Which Buddha statue has green tiles in front of it?
} \\
\midrule
\multicolumn{1}{c}{\textbf{Golden}} & \multicolumn{3}{c}{\textbf{Retrieval (Top 3)}} \\
\midrule

\raisebox{-0.5\height}{\includegraphics[width=0.23\textwidth,height=0.23\textwidth]{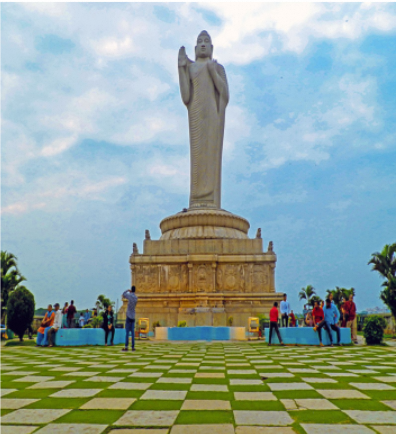}} &
\raisebox{-0.5\height}{\includegraphics[width=0.23\textwidth,height=0.23\textwidth]{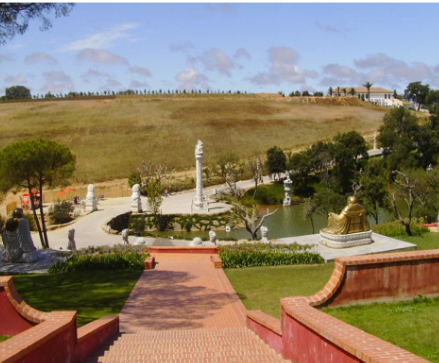}} &
\raisebox{-0.5\height}{\includegraphics[width=0.23\textwidth,height=0.23\textwidth]{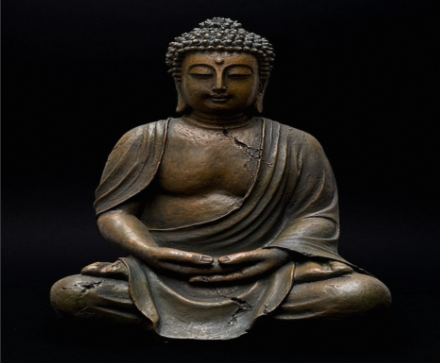}} &
\raisebox{-0.5\height}{\includegraphics[width=0.23\textwidth,height=0.23\textwidth]{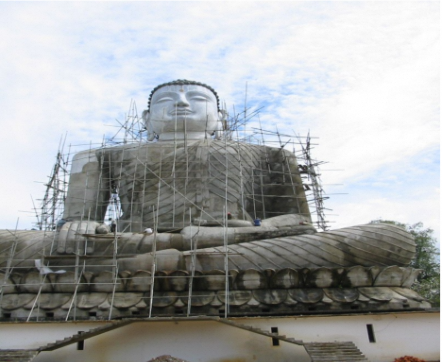}} \\

\addlinespace[1ex]  
\parbox{0.23\textwidth}{\centering Buddha statue no 1 Buddha statue at Hussain sagar } &
\parbox{0.23\textwidth}{\centering Buddha eden (34) } &
\parbox{0.23\textwidth}{\centering Buddha 1251876 } &
\parbox{0.23\textwidth}{\centering Building the Buddha – panoramio} \\
\bottomrule
\end{tabularx}

\begin{tabularx}{\textwidth}{@{}c|ccc@{}}
\toprule
\multicolumn{4}{l}{\textbf{Question:} Where can people relax in the grass next to the Eiffel Tower?
} \\
\midrule
\multicolumn{1}{c}{\textbf{Golden}} & \multicolumn{3}{c}{\textbf{Retrieval (Top 3)}} \\
\midrule

\raisebox{-0.5\height}{\includegraphics[width=0.23\textwidth,height=0.23\textwidth]{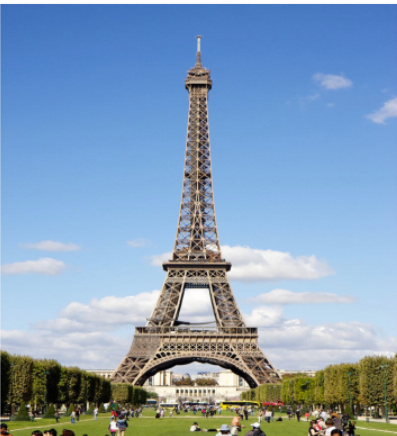}} &
\raisebox{-0.5\height}{\includegraphics[width=0.23\textwidth,height=0.23\textwidth]{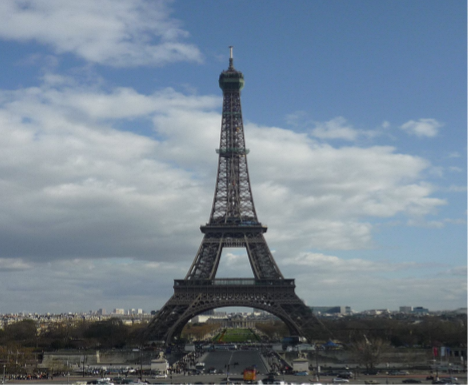}} &
\raisebox{-0.5\height}{\includegraphics[width=0.23\textwidth,height=0.23\textwidth]{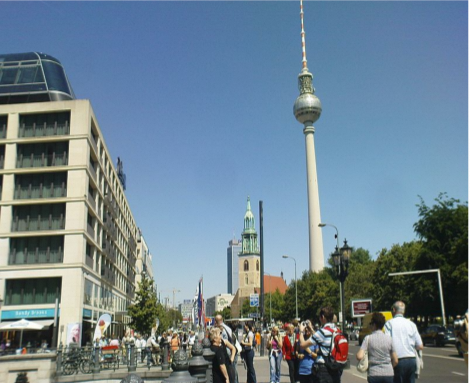}} &
\raisebox{-0.5\height}{\includegraphics[width=0.23\textwidth,height=0.23\textwidth]{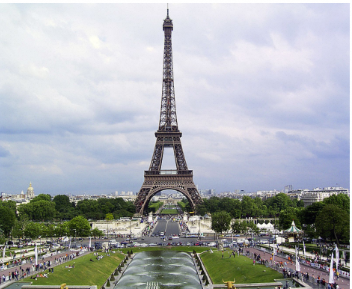}} \\

\addlinespace[1ex]  
\parbox{0.23\textwidth}{\centering Eiffel Tower, Paris 17 September 2010. } &
\parbox{0.23\textwidth}{\centering FW Eiffelturm. } &
\parbox{0.23\textwidth}{\centering Berlin , Mitte , Alexanderplatz , fernsehturm - panoramio. } &
\parbox{0.23\textwidth}{\centering Eiffelturm.} \\
\bottomrule
\end{tabularx}

\caption{Failure Case Analysis. }
\label{tab:failure}
\end{table*}
\section{The Effects of Language Models}\label{appendix:language}
The project-based approach relies on language models as the backbone, making the performance of the language model one of the key factors influencing retrieval effectiveness. To validate whether our method can function effectively with different language models as the backbone, we conduct experiments using T5-ANCE~\cite{t5ance}, BART~\cite{lewis2019bart}, BERT~\cite{lee2018pre}, GPT-2, and GPT-2-LARGE ~\cite{radford2019language}, encompassing encoder-decoder, encoder-only, and decoder-only architectures. For the encoder-decoder model, we follow MARVEL's approach\cite{zhou2024marvel} and encode the multimodal document input into the encoder, while inputting a ‘\textbackslash s’ character into the decoder segment to use its final hidden layer representation as the document representation. For the encoder-only model, we use the representation of the first character, and for the decoder-only model, we use the final hidden layer representation of the last token, as each token only computes attention with preceding tokens. Due to the large number of parameters in GPT-2 LARGE, we set its batch size to 16 to avoid memory issues, while setting the batch size to 64 for the others. Besides the results trained with the ANCE negative sampling method presented in Table \ref{tab:language_models}, we also include the results trained using the in-batch negative sampling method of DPR in Table \ref{appendixtab:language_models}. It is evident that while ANCE consistently outperforms DPR, CIEA surpasses both simple project-based and text-only methods under both negative sampling approaches, demonstrating that the proposed method aligns better with image information.

\section{Failure Case Analysis}\label{appendix:failure}
We provide a failure case analysis in Table \ref{tab:failure} for a better understanding of the proposed method and for promoting further research.
Each case presents the ground-truth image-caption pair alongside the top three retrieved candidates from \ours, with all instances classified as retrieval failures due to the absence of ground-truth matches among the top results. The first two cases illustrate the model's partial success in capturing color attributes (e.g., {\it ``orange''} and {\it ``green''}), while it might struggle to establish precise color-object mappings (e.g., {\it ``ceiling''} and {\it ``tiles''}), especially within multi-object environments. The third case reveals more fundamental limitations regarding action interpretation, as the retrieved results exhibit minimal relevance to the target action {\it ``relax''}. These failures highlight persistent challenges in visual-semantic alignment, particularly in the contextual binding of object attributes and the dynamic interpretation of action representations.

\end{document}